\documentclass[10pt, a4paper]{article}

\usepackage{lrec-coling2024} 

\usepackage{amsmath,amsfonts}
\usepackage{algorithmic}
\usepackage{algorithm}
\usepackage{array}
\usepackage{textcomp}
\usepackage{stfloats}
\usepackage{url}
\usepackage{verbatim}

\usepackage{relsize}

\usepackage{booktabs}
\usepackage{threeparttable}
\usepackage{multirow}
\usepackage{longtable}
\usepackage{adjustbox}
\usepackage{siunitx}
\usepackage{makecell}
\usepackage{tabularx}

\usepackage[group-separator={,}]{siunitx}

\usepackage{float}

\usepackage{epstopdf}
\usepackage{caption}
\usepackage{subfig}
\usepackage{xcolor}
\usepackage{pgfplots}
\usepackage{tikz}
\pgfplotsset{compat=1.10}
\usetikzlibrary{decorations.pathreplacing}
\usepackage{footmisc}

\definecolor{c1}{RGB}{47,127,193}
\definecolor{c2}{RGB}{243,210,102}
\definecolor{c3}{RGB}{232,21,24}

\DeclareMathOperator*{\argmax}{argmax}
\newcommand{\varA}[1]{{\operatorname{#1}}}

\newcommand{\ignore}[1]{}

\title{\textbf{3AM: An Ambiguity-Aware Multi-Modal Machine Translation Dataset}}

\name{Xinyu Ma$^1$, Xuebo Liu$^{2,*}$, Derek F. Wong$^{1,*}\thanks{*~Corresponding authors.}$, Jun Rao$^2$, Bei Li$^3$, \\ {\bf \large Liang Ding$^4$, Lidia S. Chao$^1$, Dacheng Tao$^4$, and Min Zhang$^2$}}

\address{
    $^1$NLP$^2$CT Lab, Department of Computer and Information Science, 
    University of Macau, Macau, China \\
    $^2$Institute of Computing and Intelligence, Harbin Institute of Technology, Shenzhen, China \\
    $^3$Northeastern University, Shenyang, China 
    $^4$The University of Sydney, Sydney, Australia \\
    \{yc27434, derekfw, lidiasc\}@um.edu.mo, \{liuxuebo, zhangmin2021\}@hit.edu.cn, \\ rao7jun@gmail.com, libei\_neu@outlook.com, liangding.liam@gmail.com, dacheng.tao@gmail.com}

\abstract{
Multimodal machine translation (MMT) is a challenging task that seeks to improve translation quality by incorporating visual information.
However, recent studies have indicated that the visual information provided by existing MMT datasets is insufficient, causing models to disregard it and overestimate their capabilities. This issue presents a significant obstacle to the development of MMT research.
This paper presents a novel solution to this issue by introducing 3AM, an ambiguity-aware MMT dataset comprising 26,000 parallel sentence pairs in English and Chinese, each with corresponding images. 
Our dataset is specifically designed to include more ambiguity and a greater variety of both captions and images than other MMT datasets. 
We utilize a word sense disambiguation model to select ambiguous data from vision-and-language datasets, resulting in a more challenging dataset.
We further benchmark several state-of-the-art MMT models on our proposed dataset. Experimental results show that MMT models trained on our dataset exhibit a greater ability to exploit visual information than those trained on other MMT datasets.
Our work provides a valuable resource for researchers in the field of multimodal learning and encourages further exploration in this area. The data, code and scripts are freely available at \url{https://github.com/MaxyLee/3AM}.
 \\ \newline  }

\begin{document}

\maketitleabstract

\begin{figure*}[t]
    \centering
    \includegraphics[width=.65\textwidth]{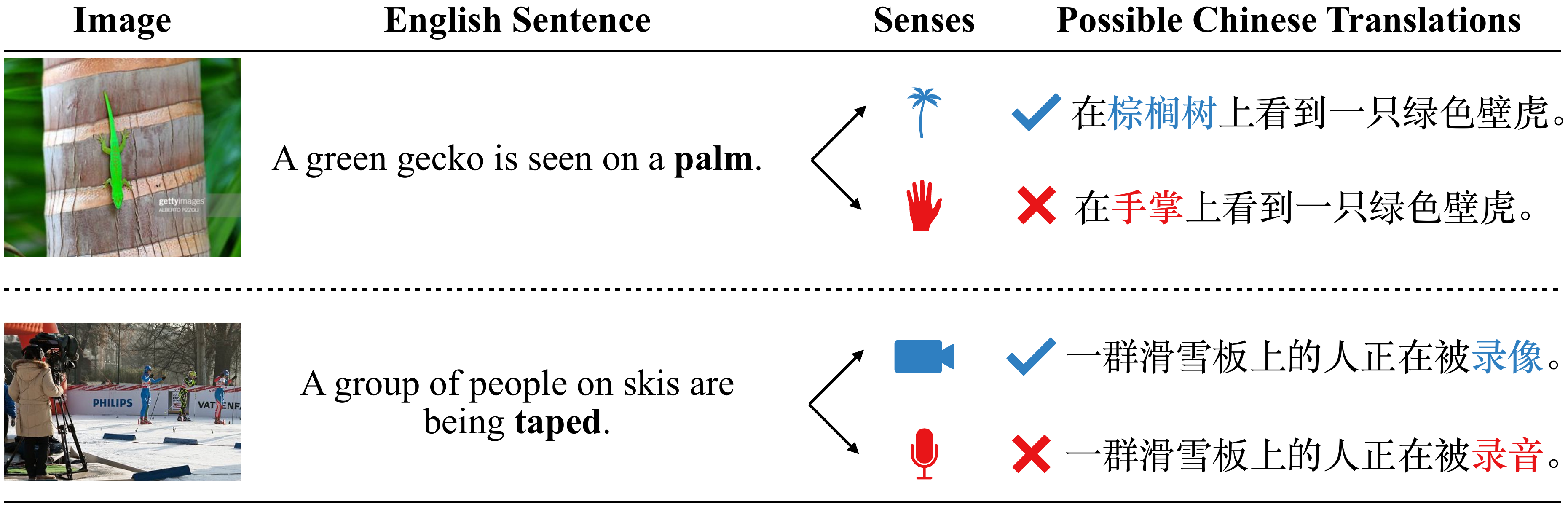}
    \caption{Examples of our 3AM dataset, where ambiguous words are shown in \textbf{bold}, with \textcolor{c3}{\bf red} and \textcolor{c1}{\bf blue} indicating its incorrect and correct translation respectively. For instance, the word `palm' has two different meanings: ``palm tree'' and ``palm of the hand''. Only from the image can it be distinguished that the correct meaning is the former.}
    \label{fig:examples}
\end{figure*}

\section{Introduction}
Multimodal machine translation (MMT) \cite{specia-etal-2016-shared} is a branch of Machine translation (MT) \cite{liu-etal-2019-shared,liu-etal-2020-norm,peng-etal-2023-towards} that involves exploiting complementary information between multiple modalities and language generation \cite{sulubacak2020multimodal}. 
MMT aims to enhance translation quality by incorporating visual information as input \cite{specia-etal-2016-shared}. Many MMT models have been proposed, which have shown superior performance compared to text-only models \cite{calixto-etal-2017-doubly, zhang2020neural, li-etal-2022-vision, zuo-etal-2023-incorporating}. However, recent studies have revealed that the visual information in existing MMT datasets contributes only marginally to translation quality. 
Experiments have shown that replacing images in the input with non-relevant images \cite{elliott-2018-adversarial} or random noise \cite{wu-etal-2021-good} has little effect on translation performance. Moreover, some studies have suggested that MMT models are less sensitive to visual information when exposed to complete sentences \cite{caglayan-etal-2019-probing, li-etal-2022-vision}. 
\citet{gronroos-etal-2018-memad, pmlr-v139-cho21a} have also indicated that pre-trained text-only models perform no worse than multimodal models in terms of translation quality.

A similar phenomenon has been observed in other multimodal tasks, such as image captioning \cite{devlin2015exploring} and VQA \cite{goyal2017making}. 
These observations suggest that natural language can provide a strong prior that can result in good superficial performance, without the underlying models truly understanding the visual content \cite{agrawal-etal-2016-analyzing, 7780911}. 
Therefore, we argue that MMT data needs to be more ambiguous and rich in visual concepts to help the models understand visual information and improve translation performance. 
As illustrated in Figure \ref{fig:examples}, given a sentence such as ``A green gecko is seen on a palm.'' the language prior is likely to lead us to interpret the word ``palm'' as meaning ``palm of the hand'' whereas the visual information suggests that it actually means ``palm tree''.

The exploitation of language priors can create the false impression that MMT models are making progress toward understanding images, when in fact, they are only leveraging language to achieve good performance. 
This can hinder progress in pushing the state-of-the-art MMT models, underscoring the urgent need for a new and more challenging MMT dataset and to better understand the circumstances under which visual information can be effectively incorporated into MMT models.
To address the limitations of existing MMT datasets, we propose 3AM, an ambiguity-aware MMT dataset that contains a larger number of ambiguous examples and a wider range of visual concepts, elevating the role of visual information understanding in MMT. Specifically, the dataset is constructed using a semi-automatic approach, which involves collecting and filtering a diverse range of data from existing vision-and-language (V+L) datasets, scoring the ambiguity of the data using a word sense disambiguation (WSD) model, and then translating the English sentences into Chinese by professional translators according to the image contents. This results in a dataset of approximately 26K pieces of data. Examples from our proposed dataset are shown in Figure \ref{fig:examples}. 

The proposed ambiguous dataset, 3AM, will compel MMT models to prioritize visual information. 
When a word is ambiguous in a sentence, the only means of resolving its meaning is by referencing the image. 
Thus, it is contended that the 3AM dataset will preclude MMT models from achieving high performance solely relying on language priors. The dataset is expected to facilitate MMT evaluation that more accurately reflects the models' ability to comprehend visual information.
To verify this hypothesis, a number of MMT models are evaluated based on 3AM and other existing MMT datasets. Extensive experiments are conducted on both text-only models and multimodal models. 
The experimental results show that the 3AM dataset presents a heightened level of complexity, necessitating the employment of models that effectively capitalize on visual patterns rather than relying solely on textual content.

Our study has three main contributions: 1) we propose an MMT dataset that is specifically designed to improve the understanding of visual information by collecting a diverse set of ambiguous sentences. Compared to existing MMT datasets, our 3AM dataset is more challenging and contains a richer set of concepts; 2) We evaluate the performance of state-of-the-art MMT models on our proposed dataset and show that models that can leverage visual information outperform text-only models. This finding supports our hypothesis that the 3AM dataset can encourage MMT models to better exploit visual information and improve translation results; 3) Our approach to constructing datasets involves collecting ambiguous data, which can also be used for other multimodal learning datasets. We hope that our work will facilitate a better understanding of the role of visual information in multimodal learning and contribute to the advancement of research.

\begin{figure*}[t]
    \centering
    \includegraphics[width=\textwidth]{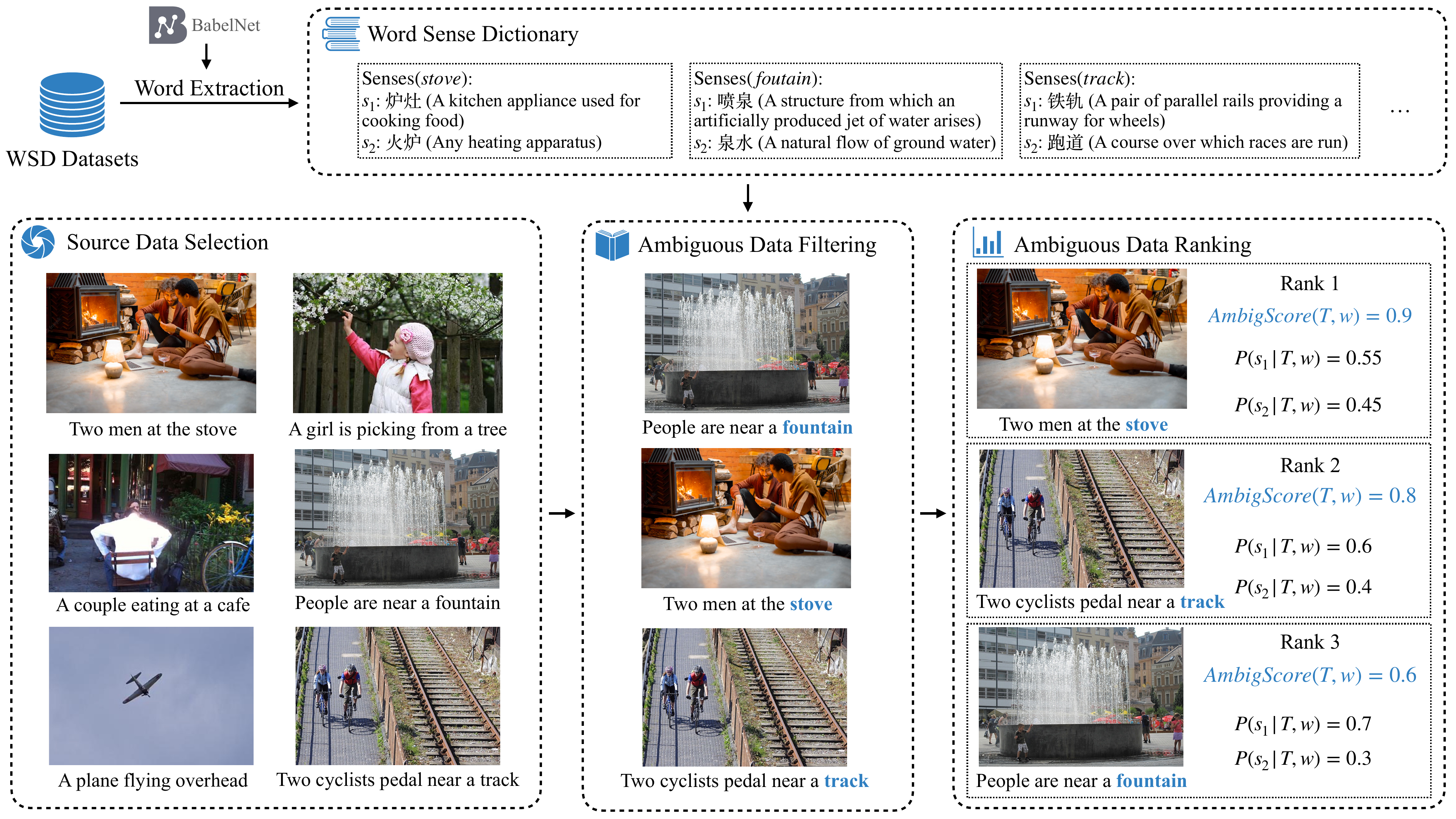}
    \caption{The process for constructing the 3AM dataset involves several steps. Firstly, we extract ambiguous words from existing WSD datasets and create a word sense dictionary using BabelNet. We then use this dictionary to filter sentences that contain ambiguous words and score them using a WSD model. Finally, we rank the sentences according to their scores to obtain the ambiguous data.}
    \label{fig:pipeline}
\end{figure*}

\section{Related Work}

MMT is a popular area of research that lies at the intersection of computer vision (CV) and natural language processing (NLP) \cite{specia-etal-2016-shared}. 
Based on the widely-used dataset Multi30K \citelanguageresource{elliott-etal-2016-multi30k}, several models were proposed to leverage visual information to improve translation performance \cite{huang-etal-2016-attention, caglayan2016multimodal, calixto-liu-2017-incorporating, delbrouck-dupont-2017-empirical, elliott-kadar-2017-imagination}. 
However, conflicting results have emerged from subsequent studies on the effectiveness of MMT models in utilizing visual input for translation \cite{elliott-2018-adversarial, wu-etal-2021-good}, raising questions about their capabilities. 
Although some work has demonstrated that models can use visual information to generate good translations under limited textual contexts such as some words in the source sentences are masked \cite{caglayan-etal-2019-probing}, the role of visual information remains unexplored when the text is complete. 
To address this issue, a number of datasets were proposed. \citetlanguageresource{li-etal-2021-vision} proposed Ambiguous Captions (AmbigCaps), a gender-ambiguous dataset in which gender information is concealed by back-translation from a gender-neutral language. 
\citetlanguageresource{li-etal-2022-visa} constructed VISA, a video-guided machine translation dataset by collecting ambiguous subtitles and corresponding video clips. 
Moreover, other datasets such as MSCTD \citelanguageresource{liang-etal-2022-msctd} and MultiSubs \citelanguageresource{wang-etal-2022-multisubs} were proposed, but these do not focus on ambiguity. 
In this work, we aim to collect more diverse and ambiguous data using automatic methods, resulting in a more challenging MMT dataset. 
Compared to previous datasets, our dataset has different source data domains and language pairs, enabling greater diversity and facilitating a better understanding of the utilization of visual features in machine translation.

\section{Dataset Construction}
\label{sec:dataset}

This section elaborates on the procedures for constructing the English-Chinese MMT dataset. 
Our primary aim is to develop an MMT dataset encompassing a wider range of visual concepts and more ambiguous data. 
To construct the dataset, we use existing V+L datasets as the source data (Sec. \ref{sec:data_src}), from which ambiguous sentences are selected. 
The data selection process can be divided into two steps: Data Filtering (Sec. \ref{sec:data_filter}) and Data Ranking (Sec. \ref{sec:data_wsd}). 
First, we construct a word sense dictionary based on the existing WSD datasets and filter the sentences accordingly to obtain data containing ambiguous words. 
Then, we use a WSD model to score and rank the data acquired in the previous step to get the most likely ambiguous sentences. 
The complete process of building the dataset is illustrated in Figure \ref{fig:pipeline}.

\begin{table}[t]
\centering
\resizebox{.48\textwidth}{!}{\begin{tabular}{lccccc}
\toprule
\textbf{Dataset} & \textbf{VE} & \textbf{COCO} & \textbf{SBU} & \textbf{CC} & \textbf{CC12M} \\
\midrule
\textbf{\# Sent.} & 570K & 616K & 1M & 3.33M & 12.4M \\
\textbf{\# Filtered Sent.} & 348K & 616K & 786K & 1.59M & 2.28M \\
\bottomrule
\end{tabular}}
\caption{Source data statistics.}
\label{tab:src_data_statistics}
\end{table}

\subsection{Data Source}
\label{sec:data_src}

For the first part, we collect the source data from five existing V+L datasets: 
\begin{enumerate}
    \item \textbf{VE}. SNLI-VE \citelanguageresource{xie2018visual} is a dataset for a classification task to determine the relationship between an image and a natural language statement. We use sentences with the label \emph{entailment}, which correspond to the content of the image, as our data source.
    \item \textbf{COCO}. COCO \citelanguageresource{lin2014microsoft} is a large-scale human-annotated multimodal dataset that contains approximately \num{123000} images and 5-way image-caption annotations.
    \item \textbf{SBU}. SBU Captions \citelanguageresource{ordonez2011im2text} is an automatically constructed dataset that contains 1 million images based on a web-scale collection of captioned images from the internet.
    \item \textbf{CC}. Conceptual Captions (CC) \citelanguageresource{sharma2018conceptual} is an image alt-text dataset that has more than 3 million images with natural-language captions. In contrast with COCO, images and captions of CC are acquired from the web, and therefore contain a wider variety of styles.
    \item \textbf{CC12M}. Conceptual 12M (CC12M) \citelanguageresource{changpinyo2021conceptual} is a web-acquired dataset with about 12 million image-text pairs that covers a more diverse set of visual concepts than CC.
\end{enumerate}

Considering that a large portion of the source data is noisy, we filter the data to ensure quality, which is divided into two steps. The first step is a rule-based filtering approach:
\begin{enumerate}
    \item \textbf{Sentence Length Filtering}. We remove sentences with fewer than 5 or more than 30 tokens, as the former are less likely to be ambiguous, while the latter tend to be poorly phrased.
    \item \textbf{URL Filtering}. As images under the same URL prefix are often of the same kind, we summarize the collection of URLs with poor image quality and filter out the corresponding data. 
    \item \textbf{Keyword Filtering}. We remove sentences that contain certain keywords, which are often of poor quality. 
    \item \textbf{Proper Noun Filtering}. Finally, sentences with more proper nouns in them are also removed.
\end{enumerate}
The second step is language model filtering, where we use a language model to score the text and filter out the poor-quality captions. Specifically, we use the GPT-2 \cite{radford2019language} model to evaluate the grammatical correctness of the captions. Given a caption $T$, the score can be formally defined as:
\begin{equation}
    \mathrm{GPTScore} = \exp(\mathrm{GPT} (T))
\end{equation}
where the $\mathrm{GPT}(\cdot)$ denotes the output of the GPT-2 model. A higher score indicates that the caption is more likely to be grammatically incorrect. After calculating the scores, we filter out the data with scores above a certain threshold based on the distribution of scores in the dataset. 
\begin{figure*}[t!]
    \centering
    \setlength{\tabcolsep}{2pt}{
        \begin{tabular}{cc}
            \makecell*[c]{
                \subfloat{
                \resizebox{0.35\textwidth}{!}{
                \begin{tikzpicture}
                    \begin{axis}[
                        ybar,
                        ymin=0,
                        width  = 12cm,
                        height = 7cm,
                        bar width=5pt,
                        ylabel={Percentage (\%)},
                        xtick = data,
                        table/header=false,
                        table/row sep=\\,
                        xticklabels from table={\footnotesize
                          0\\1\\2\\3\\4\\5\\6\\7\\8\\$\geq$9\\
                          }{[index]0},
                        enlarge y limits={value=0.05,upper}
                    ]
                    \legend{Multi30K, MSCTD, 3AM}
                    \addplot[fill=c1] table[x expr=\coordindex,y index=0]{
                        0.02\\1.02\\13.68\\31.82\\28.79\\15.30\\6.21\\2.21\\0.66\\0.29\\
                        };
                    \addplot[fill=c2] table[x expr=\coordindex,y index=0]{
                        42.30\\36.36\\15.92\\4.36\\0.90\\0.13\\0.02\\0.002\\0.00\\0.00\\
                    };
                    \addplot[fill=c3] table[x expr=\coordindex,y index=0]{
                        0.57\\3.62\\15.75\\25.18\\22.69\\15.31\\7.82\\4.30\\2.44\\2.29\\
                    };
                    \end{axis}
                \end{tikzpicture}}
                \label{fig:stat-noun}
                }
            } & 
            \makecell*[c]{
                \subfloat{
                \resizebox{0.35\textwidth}{!}{
                \begin{tikzpicture}
                    \begin{axis}[
                        ybar,
                        ymin=0,
                        width  = 12cm,
                        height = 7cm,
                        bar width=5pt,
                        ylabel={Percentage (\%)},
                        xtick = data,
                        table/header=false,
                        table/row sep=\\,
                        xticklabels from table={
                          0\\1\\2\\3\\4\\$\geq$5\\
                          }{[index]0},
                        enlarge y limits={value=0.05,upper}
                    ]
                    \legend{Multi30K, MSCTD, 3AM}
                    \addplot[fill=c1] table[x expr=\coordindex,y index=0]{
                        6.86\\55.33\\30.04\\6.65\\1.00\\0.12\\
                        };
                    \addplot[fill=c2] table[x expr=\coordindex,y index=0]{
                        29.41\\43.28\\21.24\\5.23\\0.75\\0.07\\
                    };
                    \addplot[fill=c3] table[x expr=\coordindex,y index=0]{
                        23.22\\44.48\\22.53\\7.29\\1.88\\0.58\\
                    };
                    \end{axis}
                \end{tikzpicture}}
                \label{fig:stat-verb}
                }
            } \\
            \small (a) Distributions of unique nouns per caption & \small (b) Distributions of unique verbs per caption \\
            \makecell*[c]{
                \subfloat{
                \resizebox{0.35\textwidth}{!}{
                \begin{tikzpicture}
                    \begin{axis}[
                        ybar,
                        ymin=0,
                        width  = 12cm,
                        height = 7cm,
                        bar width=2pt,
                        ylabel={Percentage (\%)},
                        xtick = data,
                        table/header=false,
                        table/row sep=\\,
                        xticklabels from table={
                          $\leq$5\\ 6\\7\\8\\9\\10\\11\\12\\13\\14\\
                          15\\16\\17\\18\\19\\20\\21\\22\\23\\24\\$\geq$25\\
                          }{[index]0},
                        enlarge y limits={value=0.05,upper}
                    ]
                    \legend{Multi30K, MSCTD, 3AM}
                    \addplot[fill=c1] table[x expr=\coordindex,y index=0]{
                        0.19\\ 0.85\\ 2.83\\ 6.02\\ 8.55\\10.50\\ 11.48\\ 10.92\\ 9.72\\ 8.61\\
                        7.20\\ 5.37\\ 4.50\\ 3.34\\ 2.67\\1.97\\ 1.48 \\ 1.03\\ 0.73\\ 0.57\\ 1.46\\
                        };
                    \addplot[fill=c2] table[x expr=\coordindex,y index=0]{
                        21.86\\ 9.99\\ 11.02\\ 11.09\\ 10.20\\ 8.98\\ 7.33\\ 6.06\\ 4.49\\ 3.34\\ 
                        2.26\\ 1.42\\ 0.86\\ 0.51\\ 0.29\\ 0.14\\ 0.09\\ 0.05\\ 0.02\\ 0.01\\ 0.00\\
                    };
                    \addplot[fill=c3] table[x expr=\coordindex,y index=0]{
                        0.41\\ 4.21\\ 5.76\\ 6.79\\ 9.35\\ 9.85\\ 9.15\\ 7.81\\ 6.82\\ 5.58\\ 
                        4.92\\ 4.30\\ 3.77\\ 3.37\\ 2.97\\ 2.78\\ 2.31\\ 2.12\\ 1.79\\ 1.62\\ 4.31\\
                    };
                    \end{axis}
                \end{tikzpicture}}
                \label{fig:stat-len}
                }
            } & 
            \makecell*[c]{
                \subfloat{
                \resizebox{0.35\textwidth}{!}{
                \begin{tikzpicture}
                \begin{axis}[
                    ybar,
                    ymin=0,
                    width  = 12cm,
                    height = 7cm,
                    bar width=2pt,
                    ylabel={Percentage (\%)},
                    xtick = {-0.5,0.5,1.5,2.5,3.5,4.5,5.5,6.5,7.5,8.5,9.5,10.5,11.5,12.5,13.5,14.5,15.5,16.5,17.5,18.5},
                    table/header=false,
                    table/row sep=\\,
                    xticklabels={0.5,,,,0.6,,,,0.7,,,,0.8,,,,0.9,,,,},
                    extra x ticks = {19.5},
                    extra x tick labels = {1.0},
                    enlarge y limits={value=0.05,upper},
                    legend pos=north west
                ]
                \legend{Multi30K, MSCTD, 3AM}
                \addplot[fill=c1] table[x expr=\coordindex,y index=0]{
                    5.04\\ 5.20\\ 4.68\\ 4.93\\ 4.88\\ 4.40\\ 4.56\\ 4.61\\ 4.65\\ 4.53\\ 
                    4.55\\ 4.92\\ 5.01\\ 5.05\\ 5.02\\ 5.30\\ 5.67\\ 5.66\\ 5.64\\ 5.72\\
                    };
                \addplot[fill=c2] table[x expr=\coordindex,y index=0]{
                    3.91\\ 4.12\\ 4.02\\ 4.28\\ 4.16\\ 4.34\\ 4.29\\ 4.40\\ 4.50\\ 4.42\\ 
                    4.83\\ 4.70\\ 5.04\\ 5.36\\ 5.48\\ 5.53\\ 6.13\\ 6.26\\ 6.86\\ 7.37\\
                };
                \addplot[fill=c3] table[x expr=\coordindex,y index=0]{
                    3.60\\ 3.39\\ 3.55\\ 3.39\\ 3.72\\ 3.59\\ 3.73\\ 3.94\\ 3.84\\ 4.22\\ 
                    4.40\\ 4.56\\ 4.97\\ 5.16\\ 5.54\\ 6.10\\ 6.67\\ 7.21\\ 8.18\\ 10.24\\
                };
                \end{axis}
            \end{tikzpicture} }
            \label{fig:stat-ambig}}
            } \\
            \small (c) Distributions of caption lengths & \small (d) Distributions of ambiguity scores
        \end{tabular}
        }
    \caption{Statistical histogram distributions on Multi30K, MSCTD, and 3AM. Compared with other datasets, 3AM contains longer captions with more unique nouns and verbs and higher ambiguity scores.}
    \label{fig:stat}
\end{figure*}

\begin{figure}[t!]
    \centering
    \resizebox{.35\textwidth}{!}{
    \begin{tikzpicture}
        \begin{axis}[
            width  = 12cm,
            height = 8cm,
            ylabel={Percentage (\%)},
            xticklabel style={rotate=90},
            xmin=0, xmax=38,
            ymin=0,
            xtick = {1,2,...,38},
            xticklabels={man, woman, wear, people, sit, stand, young, play, white, girl, dog, black, walk, boy, blue, red, child, hold, street, look, group, front, water, green, large, other, person, old, small, building, player, next, work, hand, take, table, tree},
            enlarge y limits={value=0.05,upper}
        ]
        \legend{Multi30K, 3AM}
        \addplot[color=c1,mark=*] plot coordinates {
            (1,11.40) (2,6.00) (3,3.25) (4,3.09) (5,3.06) (6,3.04) (7,2.84) (8,2.73) (9,2.66) (10,2.63) (11,2.55) (12,2.47) (13,2.47) (14,2.40) (15,2.27) (16,2.04) (17,2.02) (18,1.97) (19,1.84) (20,1.84) (21,1.82) (22,1.54) (23,1.35) (24,1.08) (25,1.06) (26,0.98) (27,0.87) (28,0.86) (29,0.85) (30,0.79) (31,0.78) (32,0.78) (33,0.73) (34,0.71) (35,0.68) (36,0.63) (37,0.61)};
        \addplot[color=c3,mark=*] plot coordinates {
            (1,6.43) (2,3.60) (3,1.31) (4,2.68) (5,2.83) (6,2.36) (7,1.81) (8,1.52) (9,2.94) (10,2.08) (11,1.68) (12,1.26) (13,1.85) (14,1.24) (15,1.59) (16,1.74) (17,1.22) (18,1.99) (19,2.09) (20,1.74) (21,1.24) (22,1.62) (23,2.08) (24,1.20) (25,1.59) (26,1.16) (27,1.85) (28,1.16) (29,1.54) (30,1.20) (31,1.31) (32,1.14) (33,1.08) (34,1.35) (35,1.62) (36,1.67) (37,1.85)
        };
        \end{axis}
    \end{tikzpicture}
    \label{fig:stat-words}
    }
    \caption{Plot of the most common words in the captions of Multi30K and 3AM, the words in the 3AM dataset are more evenly distributed.}
    \label{fig:stat2}
\end{figure}
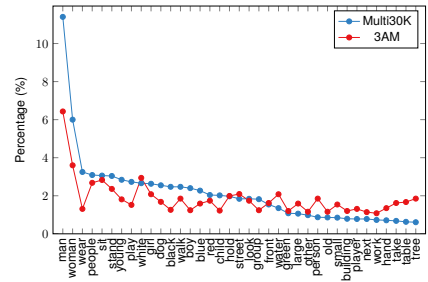

\subsection{Data Selection}
\label{sec:data_select}

\subsubsection{Data Filtering}
\label{sec:data_filter}

The first step is to construct a word sense dictionary using the existing WSD datasets based on which the data containing ambiguous words can be obtained. The construction of the word sense dictionary consists of the following steps:
\begin{enumerate}
    \item \textbf{Vocabulary Collection}. 
    We collect ambiguous words from the following WSD datasets to construct the ambiguous vocabulary: MuCoW \citelanguageresource{raganato-etal-2020-evaluation}, OneSeC \citelanguageresource{scarlini-etal-2019-just}, Train-o-Matic \citelanguageresource{pasini-navigli-2017-train}, SemCor \citelanguageresource{miller-etal-1994-using}, and OMSTI \citelanguageresource{taghipour-ng-2015-one}. 
    The definitions of each word in these datasets along with the WordNet \citelanguageresource{miller-1994-wordnet} sense keys are saved for word sense disambiguation. 
    Since a word may have different senses for different parts of speech (POS), we add words to the vocabulary with their POS as well following \citetlanguageresource{li-etal-2022-visa}. For example, for the noun `fountain', we add `fountain.n' to the vocabulary.
    \item \textbf{Word Sense Dictionary}. 
    In this part, we build a dictionary containing ambiguous words with their Chinese translations and corresponding definitions. 
    For each word in the vocabulary, we use BabelNet \cite{navigli2012babelnet}, a large-scale multilingual encyclopedic dictionary, to get its Chinese translations and the corresponding senses.
    Then, we filter out words with only one sense based on the word sense dictionary, which is considered not ambiguous. 
    As illustrated in Figure \ref{fig:pipeline}, given a word $w$, we can get the set of its senses $S=\mathrm{Senses}(w)$ with its Chinese translation from the word sense dictionary. 
\end{enumerate}

After constructing the word sense dictionary, we can use it to select sentences containing ambiguous words. 
First, we use spaCy \cite{spacy2} to tokenize the sentences, convert words to the base form, and get the POS of words. 
Then we match these sentences with the dictionary to get data containing ambiguous words. 
Note that a sentence may contain more than one ambiguous word, they are considered as different data entries in the next step.

\subsubsection{Word Sense Disambiguation}
\label{sec:data_wsd}

As the second step, we apply an automatic method to determine whether words are ambiguous in context. We use a WSD model (ConSeC) \cite{barba-etal-2021-consec} to score the data obtained in the previous step. 
Given a sentence $T$ and a target word $w$, the probability distribution $P$ of the set of senses $S$ of the target word in the context can be defined as:
\begin{equation}
    P(S|T, w) = \mathrm{WSD}(T, w, S)
\end{equation}
where the $\mathrm{WSD}(\cdot)$ is the output of the WSD model. Then, following \citetlanguageresource{pasini-navigli-2017-train}, the ambiguity score of word $w$ in sentence $T$ can be defined as:
\begin{equation}
    \mathrm{AmbigScore}(T, w) = P(s_{1}|T, w) - P(s_{2}|T, w)
\end{equation}
where $s_{1} = \argmax_{s \in S} P(s|T, w)$ and $s_{2} = \argmax_{s \in S \backslash \{s_{1}\}} P(s|T, w)$. 
We then sort the data using the ambiguity score in descending order and get a ranked list of data that the WSD model considers most likely to be ambiguous.
To ensure the diversity of ambiguous words, we set an upper bound on the number of occurrences of each sense and get about \num{30000} pieces of data for annotation. 

\subsection{Annotation}
\label{sec:data_annot}

We engaged professional annotators from translation companies to translate our data. 
If the sentence does not match the image or is of low quality, the annotators will remove it. 
To ensure high quality, the annotated data are sampled in batches, and if the error rate is high, the entire batch is checked. A total of approximately \num{26000} items of data are annotated, and we manually select \num{1000} highly ambiguous sentences for inclusion in our validation and test sets.

\subsection{Dataset Statistics}
\label{sec:data_stat}

 \begin{table*}[t]
\centering
\small
\begin{tabular}{lrccccccc}
\toprule
\multirow{2}{*}{\textbf{Dataset}} & \multicolumn{5}{c}{\textbf{Text}} & \multicolumn{3}{c}{\textbf{Image}} \\
\cmidrule(lr){2-6} \cmidrule(lr){7-9}
& \textbf{Avg. length} & \textbf{Dist-1} & \textbf{Dist-2} & \textbf{Dist-3} & \textbf{Dist-4} & \textbf{LPIPS} & \textbf{IS} & \textbf{Ent-Obj} \\
\midrule
Multi30K & 13.06 & 0.25 & 2.29 & 5.26 & 7.31 & 0.80584 $\pm$ 0.00010 & 23.25 $\pm$ 2.58 & 3.15 \\
MSCTD & \: 8.40 & 0.17 & 1.38 & 3.16 & 4.07 & 0.74149 $\pm$ 0.00011 & \: 7.85\ $\pm$ 0.20 & 3.21 \\
3AM & 13.48 & 0.77 & 5.23 & 8.85 & 9.67 & 0.82975 $\pm$ 0.00011 & 29.94 $\pm$ 3.75 & 4.35 \\
\bottomrule
\end{tabular}
\caption{Detailed statistics of Multi30K, MSCTD, and 3AM. We demonstrate the average sentence length, the LPIPS (higher means more different), the average number of unique n-grams, and the average number of unique words. Our 3AM dataset is lexically richer than other datasets in general.}
\label{tab:dataset_stat}
\end{table*}

\subsubsection{Basic Statistics}
We conduct a comprehensive analysis of the 3AM dataset and other MMT datasets. 
The statistical properties of Multi30K, MSCTD, and our proposed 3AM are shown in Figure \ref{fig:stat}. 
We calculate the number of unique words (noun and verb) in each sentence across different datasets. 
As shown in Figure \ref{fig:stat-noun} and \ref{fig:stat-verb}, 3AM contains more unique words than other datasets, showing that 3AM is of higher lexical richness.
In addition, Figure \ref{fig:stat-len} illustrates that the captions of 3AM are longer than others. 
We also use a WSD model to score the ambiguity of the data. 
It can be seen from Figure \ref{fig:stat-ambig} that the 3AM dataset has a higher ambiguity. 
Besides, we count the most common words and, as shown in Figure \ref{fig:stat2}, the words in our dataset are more evenly distributed. 
Here we do not compare with MSCTD as its lexical distribution differs considerably from the other datasets.

\subsubsection{Diversity Analysis}
Additionally, we conduct a quantitative analysis of diversity, which is presented in Table \ref{tab:dataset_stat}. 
To evaluate text diversity, we calculate the average sentence length and the proportion of the number of distinct n-grams to the total number of n-grams (Dist-n) following \citet{li-etal-2016-diversity}. 
The results show that 3AM contains longer captions with a greater number of unique n-grams, demonstrating that it is more lexically rich than other datasets. 
To assess the diversity of images, we use the Learned Perceptual Image Patch Similarity (LPIPS) \cite{zhang2018perceptual} and Inception Score (IS) \cite{salimans2016improved}. 
For LPIPS, we randomly sample \num{1000} images from each dataset and calculate the perceptual distance between each pair of images following \citet{lee2018diverse}. 
As shown in Table \ref{tab:dataset_stat}, the LPIPS and IS of 3AM is significantly higher, indicating greater image diversity. 
Moreover, to evaluate the diversity of objects in the images, we propose to use the Entropy of Objects (Ent-Obj) following \citet{zhang2018generating}:
\begin{equation}
    \varA{Ent-Obj} = -\frac{1}{\sum_{o}F(o)}\sum_{o \in O}F(o)\log(\frac{F(o)}{\sum_{o}F(o)})
\end{equation}
where $O$ is the set of all objects, and $F(o)$ denotes the frequency of object $o$. 
Specifically, we use YOLOv6 \cite{li2022yolov6} to detect objects in the images. 
The results show that, unlike other datasets where the majority of objects are `persons', etc., the objects in 3AM are more evenly distributed.

\section{Experiments}
\label{sec:exp}

\subsection{Datasets}
\label{sec:exp_dataset}

For the dataset, in addition to experiments on 3AM, we also conduct experiments on other MMT datasets for comparison.

\subsubsection{Multi30K}

Multi30K \citelanguageresource{elliott-etal-2016-multi30k} is a commonly used MMT dataset based on Flickr30K \citelanguageresource{young2014image} with the manual translation of English captions into German and French with approximately \num{30000} pieces of data, including three testsets: test2016, test2017, and mscoco. Here we use test2017 as the test set.
For comparison with 3AM, we create an English-Chinese version of Multi30K by translating the English sentences into Chinese by a strong English-Chinese MT model.

\subsubsection{MSCTD}

MSCTD \citelanguageresource{liang-etal-2022-msctd} is a multimodal sentiment chat translation dataset that contains \num{142871} English-Chinese parallel sentences and \num{30370} English-German parallel sentences. 
We use English-Chinese translation data from MSCTD to compare with our dataset. 

\begin{table*}[t]
\centering
\small
\begin{tabular}{rcccccccccccc}
\toprule
\multirow{4}{*}{\bf Method} & \multicolumn{12}{c}{\textbf{Multi30K (train)}} \\
\cmidrule(lr){2-13}
& \multicolumn{4}{c}{Multi30K (test)} & \multicolumn{4}{c}{MSCTD (test)} & \multicolumn{4}{c}{3AM (test)} \\
\cmidrule(lr){2-5} \cmidrule(lr){6-9} \cmidrule(lr){10-13}
& B $\uparrow$ & BS $\uparrow$ & M $\uparrow$ & T $\downarrow$ & B $\uparrow$ & BS $\uparrow$ & M $\uparrow$ & T $\downarrow$ & B $\uparrow$ & BS $\uparrow$ & M $\uparrow$ & T $\downarrow$ \\
\midrule
Trans & 42.86 & 74.32 & 65.44 & 47.86 & 2.87 & 34.99 & 15.75 & 108.20 & 10.86 & 49.10 & 29.40 & 88.85 \\
SelAttn & 42.00 & 74.17 & 64.63 & 49.82 & 2.86 & 36.00 & 16.61 & 107.84 & 11.67 & 50.05 & 30.86 & 87.20 \\
\cmidrule(lr){1-13}
Bart & 56.93 & 83.24 & 79.61 & 32.47 & 7.40 & 46.71 & 29.35 & 101.93 & 22.29 & 59.19 & 45.43 & 73.87 \\
VL-Bart & 56.70 & 82.93 & 77.89 & 32.00 & 8.12 & 46.29 & 27.22 & \: 86.40 & 23.20 & 60.20 & 45.75 & 70.95 \\
\cmidrule(lr){1-13}
T5 & \textbf{60.59} & \textbf{85.69} & \textbf{82.85} & \textbf{27.61} & 10.24 & 52.53 & \textbf{38.78} & \: 85.30 & 25.03 & 62.99 & 50.72 & 67.08 \\
VL-T5 & 59.61 & 85.25 & 82.12 & 27.95 & \textbf{11.10} & \textbf{52.96} & 38.71 & \: \textbf{77.71} & \textbf{25.34} & \textbf{63.25} & \textbf{50.89} & \textbf{66.35} \\ 
\midrule
\midrule
\multirow{4}{*}{\bf Method} & \multicolumn{12}{c}{\textbf{MSCTD (train)}}\\ 
\cmidrule(lr){2-13}
& \multicolumn{4}{c}{Multi30K (test)} & \multicolumn{4}{c}{MSCTD (test)} & \multicolumn{4}{c}{3AM (test)} \\
\cmidrule(lr){2-5} \cmidrule(lr){6-9} \cmidrule(lr){10-13}
& B $\uparrow$ & BS $\uparrow$ & M $\uparrow$ & T $\downarrow$ & B $\uparrow$ & BS $\uparrow$ & M $\uparrow$ & T $\downarrow$ & B $\uparrow$ & BS $\uparrow$ & M $\uparrow$ & T $\downarrow$ \\
\midrule
Trans & \: 9.89 & 50.43 & 30.75 & 80.68 & 22.97 & 62.93 & 46.43 & 65.40 & 4.51 & 40.69 & 20.10 & \: 88.37 \\
SelAttn & \: 6.91 & 46.75 & 25.04 & 85.31 & 20.87 & 62.08 & 44.27 & 65.58 & 5.30 & 41.87 & 21.05 & 108.70 \\
\cmidrule(lr){1-13}
Bart & 22.77 & 65.66 & 51.50 & 59.95 & \textbf{32.68} & 69.82 & \textbf{56.68} & \textbf{52.60} & 14.93 & 56.34 & 38.72 & \: 74.58 \\
VL-Bart & 18.10 & 60.34 & 44.81 & 65.29 & 30.81 & 68.96 & 55.63 & 54.03 & 13.61 & 54.24 & 36.46 & \: 77.53 \\
\cmidrule(lr){1-13}
T5 & \textbf{29.17} & \textbf{72.04} & \textbf{59.82} & \textbf{51.32} & 29.39 & 70.43 & 54.22 & 54.46 & \textbf{18.49} & \textbf{59.68} & \textbf{44.13} & \: \textbf{70.26} \\
VL-T5 & 28.43 & 71.09 & 58.85 & 52.82 & 29.49 & \textbf{70.63} & 54.48 & 54.52 & 17.87 & 59.27 & 43.44 & \: 70.55 \\
\midrule
\midrule
\multirow{4}{*}{\bf Method} & \multicolumn{12}{c}{\textbf{3AM (train)}} \\
\cmidrule(lr){2-13}
& \multicolumn{4}{c}{Multi30K (test)} & \multicolumn{4}{c}{MSCTD (test)} & \multicolumn{4}{c}{3AM (test)} \\
\cmidrule(lr){2-5} \cmidrule(lr){6-9} \cmidrule(lr){10-13}
& B $\uparrow$ & BS $\uparrow$ & M $\uparrow$ & T $\downarrow$ & B $\uparrow$ & BS $\uparrow$ & M $\uparrow$ & T $\downarrow$ & B $\uparrow$ & BS $\uparrow$ & M $\uparrow$ & T $\downarrow$ \\
\midrule
Trans & 25.95 & 64.51 & 49.88 & 63.92 & 3.53 & 39.23 & 19.02 & 102.93 & 11.33 & 49.51 & 31.34 & 89.68 \\
SelAttn & 27.81 & 67.06 & 52.13 & 59.77 & 4.25 & 40.34 & 19.84 & 100.19 & 13.33 & 51.54 & 33.47 & 87.05 \\
\cmidrule(lr){1-13}
Bart & 48.13 & 80.16 & 76.07 & 39.19 & 13.45 & 54.61 & 38.30 & \: 84.94 & 31.47 & 65.87 & 55.62 & 63.65 \\
VL-Bart & 50.13 & 80.74 & 76.38 & 36.87 & 16.13 & 56.45 & 39.15 & \: 74.17 & 33.27 & 66.56 & 55.84 & 61.28 \\
\cmidrule(lr){1-13}
T5 & 50.16 & 81.84 & 79.18 & 35.92 & 15.56 & 59.18 & 48.04 & \: 77.79 & 33.09 & 68.15 & 57.26 & 60.09  \\
VL-T5 & \textbf{52.04} & \textbf{82.60} & \textbf{79.76} & \textbf{34.37} & \textbf{17.12} & \textbf{59.94} & \textbf{48.54} & \: \textbf{73.01} & \textbf{34.24}& \textbf{68.39} & \textbf{59.12} & \textbf{58.88}  \\
\bottomrule
\end{tabular}
\caption{Performance of MMT models on 3AM and other MMT datasets in terms of BLEU (B), BERT-Score (BS), METEOR (M), and TER (T). }
\label{tab:main_result}
\end{table*}

\subsection{Baseline Models}
\label{sec:exp_models}

In order to provide a convincing benchmark for 3AM, we conduct experiments on multiple machine translation models as the baseline models, which can be divided into two categories: Text-only models and Multimodal models. 

\subsubsection{Text-only models}

\begin{enumerate}
    \item \textbf{Trans}. 
    We use the Transformer (Trans) \cite{NIPS2017_3f5ee243} model as one of the baseline models. 
    We conduct experiments on the Transformer-Tiny configuration following previous work \cite{wu-etal-2021-good, li-etal-2022-vision} for a fair comparison with the Selective Attention model.
    \item \textbf{Bart}. 
    Bart \cite{lewis-etal-2020-bart} is a Transformer-based model pre-trained by corrupting text with an arbitrary noise and allows the model to reconstruct it. 
    As Chinese is not included in the vocabulary of Bart, we extend it by incorporating Chinese tokens and then extending the token embeddings of the model to the corresponding dimension.
    \item \textbf{T5}. 
    T5 \cite{raffel2020exploring} is a Transformer model pre-trained on a mixture of supervised and unsupervised tasks which are converted into the text-to-text format. 
    We extend T5 using the same approach as Bart.
\end{enumerate}

\subsubsection{Multimodal models}

\begin{enumerate}
    \item \textbf{SelAttn}. 
    Selective Attention (SelAttn) \cite{li-etal-2022-vision} is an MMT model that uses vision features extracted by Transformer-based models such as Vision Transformer (ViT) \cite{dosovitskiy2021an}. 
    As previously mentioned, we use a small model size following \citet{li-etal-2022-vision}.
    \item \textbf{VL-Bart}. 
    VL-Bart \cite{pmlr-v139-cho21a} is a multimodal pre-trained model that extends the Bart model encoder to a multimodal encoder by incorporating image features as additional input. 
    We employ the same extending approach as we did for Bart and T5. 
    \item \textbf{VL-T5}. 
    Similarly, VL-T5 \cite{pmlr-v139-cho21a} is a multimodal pre-trained model extended from T5 using the same method as that of VL-Bart. 
\end{enumerate}

\subsection{Implementation Details}
\subsubsection{Training}
Our models are implemented based on PyTorch \cite{pytorch}, Fairseq \cite{ott-etal-2019-fairseq}, and Huggingface Transformers \cite{wolf-etal-2020-transformers}. 
All experiments are run on Nvidia A40 GPUs. 
The implementation details of different models are as follows:

\begin{enumerate}
    \item Models trained from scratch. 
    The Transformer and Selective Attention models are trained from scratch, consisting of 4 encoder and decoder layers. 
    We set the hidden size as 128, the filter size of FFN as 256, and 4 heads in the multi-head self-attention. 
    We use the Adam Optimizer \cite{DBLP:journals/corr/KingmaB14} with $\beta_1=.9$, $\beta_2=.98$, and $\varepsilon=10^{-8}$. 
    We train the models with a batch size of \num{4096} tokens and a learning rate of $6 \times 10^{-4}$. 
    We adopt the early stopping training strategy to prevent overfitting following \citet{li-etal-2022-vision}.
    \item Pre-trained models. 
    We use the pre-trained T5-base\footnote{https://huggingface.co/t5-base} and Bart-base\footnote{https://huggingface.co/facebook/bart-base} models from Huggingface Transformers for fine-tuning following \citet{pmlr-v139-cho21a}. 
    The models are fine-tuned with a batch size of 60 and a learning rate of $5 \times 10^{-4}$. 
    We use AdamW Optimizer \cite{loshchilov2018decoupled} with a weight decay of 0.01. 
\end{enumerate}

\subsubsection{Evaluation}
\label{sec:exp_metrics}
The model with the lowest validation perplexity is used for model testing.
The beam size for inference is set to 5.
We use BLEU \cite{papineni-etal-2002-bleu}, BERT-Score \cite{Zhang*2020BERTScore:}, METEOR \cite{banerjee-lavie-2005-meteor}, and TER \cite{snover-etal-2006-study} as the automatic evaluation metrics. 
We use SacreBLEU \cite{post2018call} to calculate the BLEU score following \citet{pmlr-v139-cho21a} and the Huggingface Transformers \cite{wolf-etal-2020-transformers} for the METEOR score and BERT-Score. 
Specifically, we use the BERT-Large-Chinese\footnote{https://huggingface.co/yechen/bert-large-chinese} model for calculating the BERT-Score.

\subsection{Results and Discussion}
\label{sec:exp_results}
The experiment results are shown in Table \ref{tab:main_result}. Each model is trained on three datasets and the performance is evaluated on each test set. 
Although MMT models trained on Multi30K perform close to or even worse than their text-only counterparts on both Multi30K and MSCTD test sets, they perform slightly better on the 3AM test set. 
This result confirms our view that although the current MMT models can exploit visual information, this ability is hardly reflected in the existing datasets as visual information plays a minor role. 
In addition, the poor performance of the MMT models trained on the MSCTD dataset compared to the text-only models illustrates the inability of MSCTD to allow the model to learn how to utilize visual information. 
Besides, models trained on the MSCTD dataset perform poorly on the other test sets as it is a dialogue dataset and differs significantly from the other data.
Moreover, MMT models trained on the 3AM dataset outperform the text-only models by a large margin, demonstrating that visual information plays a vital role in our 3AM dataset. 
These findings confirm our hypothesis that our proposed dataset will force the MMT models to leverage visual information to disambiguate and thus generate higher-quality translations.

\subsection{Analysis}
\label{sec:exp_analysis}

\begin{table}[t]
\centering
\small
\begin{tabular}{lccc}
\toprule
\textbf{Dataset} & \textbf{C} & \textbf{I} & \textbf{$\Delta$-Awareness}\\
\midrule
Multi30K & 74.16 & 74.11 $\pm$ 0.04 & 0.05 $\pm$ 0.04 \\
MSCTD & 62.08 & 62.08 $\pm$ 0.00 & 0.00 $\pm$ 0.00 \\
3AM & 51.54 & 50.17 $\pm$ 0.09 & 1.36 $\pm$ 0.09 \\
\bottomrule
\end{tabular}
\caption{BERT-Scores under Congruent (\textbf{C}) and Incongruent (\textbf{I}) settings, and the image awareness results. The Incongruent and $\Delta$-Awareness scores are calculated by taking the mean and standard deviation of the scores obtained from five random permutations of the visual data.}
\label{tab:image_awareness}
\end{table}

\begin{figure}[t]
    \centering
    \includegraphics[width=.48\textwidth]{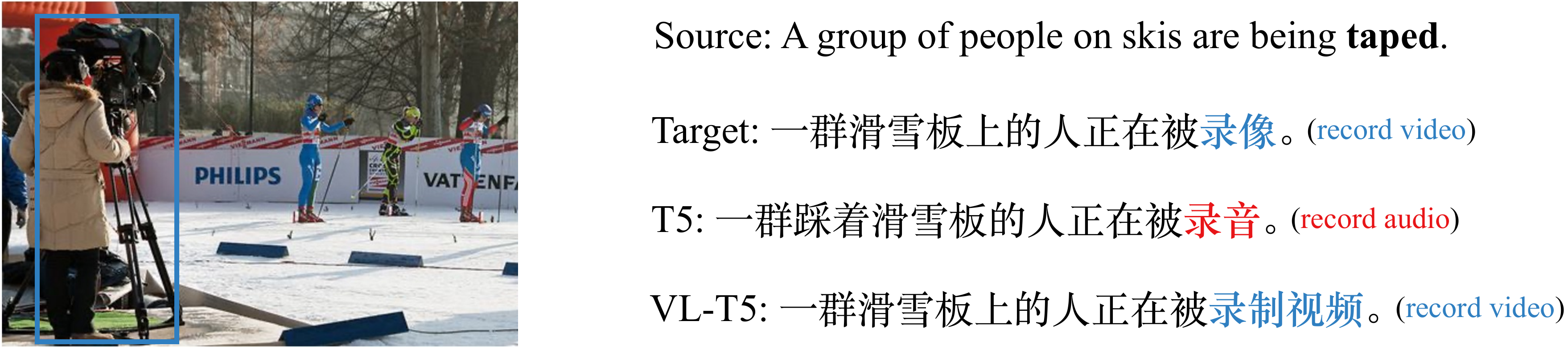}
    \caption{A case study of the 3AM dataset. The ambiguous word in the source sentence is in \textbf{bold}. The \textcolor{c3}{red} and \textcolor{c1}{blue} represent incorrectly and correctly translated words respectively.}
    \label{fig:case}
\end{figure}

\subsubsection{Visual Awareness}

To evaluate the effect that visual information has in translation, we use an adversarial evaluation method following \citet{elliott-2018-adversarial} to calculate image awareness. 
Let $x$ denote the source sentence, $y$ denote the target sentence, $v$ denote the congruent image, and $\bar{v}$ denote the incongruent image. 
The overall image awareness of a model $\mathcal{M}$ on dataset $\mathcal{D}$ can be defined as:
\begin{equation}
    \varA{\Delta-Awareness} = \frac{1}{|\mathcal{D}|}\sum^{|\mathcal{D}|}_{i}a_{\mathcal{M}}(x_i, y_i, v_i, \bar{v}_i)
\end{equation}
where the $a_{\mathcal{M}}(\cdot)$ is the image awareness of model $\mathcal{M}$ on a single instance:
\begin{equation} \label{eq:awareness}
    a_{\mathcal{M}}(x_i, y_i, v_i, \bar{v}_i) = \mathlarger{\varepsilon}(x_i, y_i, v_i) - \mathlarger{\varepsilon}(x_i, y_i, \bar{v}_i)
\end{equation}
where the $\mathlarger{\varepsilon}$ is an evaluable performance measure, and here we use the BERT-Score.

To determine whether a model passes the evaluation, we conduct a Wilcoxon signed-rank test using the pairs of BERT-scores calculated in the process of computing the image awareness and combine 5 separate $p$ values from each test using Fisher's method following \citet{elliott-2018-adversarial}.
Table \ref{tab:image_awareness} shows the evaluation results of the BERT-Scores and the image awareness scores. 
We find that while the model did not pass the evaluation on both Multi30K ($\chi ^2=15.48$, $p=0.1156$) and MSCTD ($\chi ^2=9.54$, $p=0.4819$), the visual information improves the performance on the 3AM dataset ($\chi ^2=249.46$, $p<0.0001$) by a large margin. 
This demonstrates that MMT models trained on 3AM utilize visual information to generate better translation. 

\subsubsection{Case Study}

As shown in Figure \ref{fig:case}, the word `tape' in the source sentence has two senses in Chinese: `record audio' and `record video'. 
Although it is difficult to distinguish from the text, from the image we can easily see that the real meaning is the latter. 
In this example, the T5 model incorrectly translates the word, whereas VL-T5 produces the correct translation as it can utilize visual information to disambiguate. 
This observation further confirms our hypothesis that MMT models trained on the 3AM dataset can effectively exploit visual information. 

\section{Conclusion}
\label{sec:conclusion}
In this paper, we present 3AM, a challenging MMT dataset that contains more ambiguity and diverse visual concepts, consisting of approximately 26K data points
Several MMT models are benchmarked on the proposed dataset and compared with existing MMT datasets. It is found that models with visual input outperform those without visual input, confirming the hypothesis that the ambiguous dataset forces the models to focus on visual information. This enables a more realistic evaluation of the models' performance and demonstrates the effectiveness of using visual information rather than relying solely on language priors.
The release of the 3AM dataset is expected to facilitate the advancement of research on multimodal learning.

\section{Acknowledgements}

This work was supported in part by the Science and Technology Development Fund, Macau SAR (Grant Nos. FDCT/060/2022/AFJ, FDCT/0070/2022/AMJ), National Natural Science Foundation of China (Grant No. 62261160648), Ministry of Science and Technology of China (Grant No.~2022YFE0204900), the Multi-year Research Grant from the University of Macau (Grant~No.~MYRG-GRG2023-00006-FST-UMDF), Shenzhen College Stability Support Plan (Grant Nos.~GXWD20220811173340003, GXWD20220817123150002), and Shenzhen Science and Technology Program (Grant Nos.~RCBS20221008093121053, ZDSYS20230626091203008).~Xuebo Liu was sponsored by CCF-Tencent Rhino-Bird Open Research Fund. 
This work was performed in part at SICC which is supported by SKL-IOTSC, and HPCC supported by ICTO of the University of Macau.

\section{Ethical Considerations}

Our 3AM dataset was developed on the basis of existing datasets and therefore adheres to the corresponding copyrights. We hired a professional translation company for the data annotation, a contract was signed and the staff were duly remunerated.

\section{Bibliographical References}\label{sec:reference}

\bibliographystyle{lrec-coling2024-natbib}
\bibliography{refs}

\section{Language Resource References}
\label{lr:ref}
\bibliographystylelanguageresource{lrec-coling2024-natbib}
\bibliographylanguageresource{languageresource}

\end{document}